\documentclass[pmlr]{jmlr}

\RequirePackage{graphicx}
\usepackage{booktabs}
\usepackage{longtable}
\usepackage{multirow}
\usepackage{caption}

\makeatletter
\def\set@curr@file#1{\def\@curr@file{#1}}
\makeatother
\usepackage[load-configurations=version-1]{siunitx}

\jmlrvolume{Preprint}
\jmlryear{2026}
\jmlrworkshop{}
\jmlrproceedings{Preprint}{}

\title[Good Rankings, Wrong Probabilities]{Good Rankings, Wrong Probabilities: A Calibration Audit of Multimodal Cancer Survival Models}

\author{\Name{Sajad Ghawami}
       \Email{sajad@ghawami.io}\\
       \addr Independent Researcher}

\begin{document}

\maketitle

\begin{abstract}
Multimodal deep learning models that fuse whole-slide histopathology images with genomic data have achieved strong discriminative performance for cancer survival prediction, as measured by the concordance index. Yet whether the survival probabilities derived from these models -either directly from native outputs or via standard post-hoc reconstruction -are calibrated remains largely unexamined. We conduct, to our knowledge, the first systematic fold-level 1-calibration audit of multimodal WSI-genomics survival architectures. In Experiment A, we train and directly evaluate native discrete-time survival outputs from 3 models on TCGA-BRCA: all three fail 1-calibration on a majority of folds (12 of 15 fold-level tests reject after Benjamini-Hochberg correction), providing direct evidence of miscalibration. In Experiment B, we extend the audit to 11 architectures across 5 TCGA cancer types using Breslow-reconstructed survival curves from scalar risk scores; because this reconstruction imposes a proportional-hazards assumption, the observed miscalibration may partly reflect the reconstruction step rather than the models alone. Across both experiments (290 fold-level tests), 166 reject the null of correct calibration at the median event time after Benjamini-Hochberg correction (FDR = 0.05). Models with the highest discrimination are among the most miscalibrated at this horizon: MCAT achieves C-index 0.817 on GBMLGG yet fails 1-calibration on all five folds. Gating-based fusion is associated with better calibration; bilinear and concatenation fusion are not. Post-hoc Platt scaling reduces miscalibration at the evaluated horizon (e.g., MCAT: 5/5 folds failing to 2/5) without affecting discrimination. The concordance index alone is insufficient for evaluating survival models intended for clinical use.
\end{abstract}

\section{Introduction}

Multimodal deep learning models that fuse whole-slide histopathology images with genomic profiles now achieve concordance indices of 0.60-0.82 for cancer survival prediction. Models such as MCAT \citep{chen2021multimodal}, SurvPath \citep{jaume2024modeling}, and MMP \citep{song2024multimodal} represent genuine advances, with potential for clinical risk stratification.

But there is a gap between ranking patients and advising them. The concordance index measures whether a model correctly orders patients by risk. It says nothing about the probabilities these models assign. A model can rank patients perfectly while producing survival estimates that bear little resemblance to reality. When a model outputs ``30\% five-year survival,'' that number may reach a clinician, a tumor board, or a patient. If it should have been 60\%, the consequences are concrete. In one simulation study, a confidence-calibration framework reduced clinician override rates to as low as 1.7\% for high-confidence predictions \citep{yu2025enhancing}, suggesting that well-calibrated AI outputs may strongly influence clinical decisions. If those outputs are miscalibrated, the influence persists but the guidance is wrong.

Calibration -the alignment between predicted probabilities and observed outcomes -bridges the gap between a useful ranking and a trustworthy prediction. Tools exist: D-calibration and 1-calibration \citep{haider2020effective} test whether predicted survival distributions match observed event rates, implemented in SurvivalEVAL \citep{qi2023survivaleval}. Yet across the leading multimodal WSI-genomics survival models, formal calibration has received little attention. MCAT, SurvPath, MMP, PORPOISE \citep{chen2022pan}, and MOTCat \citep{xu2023multimodal} all report only C-index. MultiSurv \citep{valesilva2021multisurv} reports IBS -a composite mixing discrimination and calibration. MADSurv \citep{zhang2025madsurv} reports Brier scores. VLSA \citep{liu2025vlsa} reports D-calibration for vision-only and vision-language models. But no prior work, to our knowledge, has conducted a systematic fold-level 1-calibration audit with post-hoc recalibration analysis for the MCAT/SurvPath/MMP family.

This reflects a broader pattern: a survey of 2023-2025 survival analysis publications found concordance metrics used 10$\times$ more frequently than calibration metrics \citep{lillelund2025stop}. The FDA's January 2025 draft guidance on AI-enabled device software functions \citep{fda2025aidsf} discusses calibration and uncertainty among considerations for performance evaluation of higher-risk applications. The gap between what regulators expect and what the field reports is wide.

We set out to close that gap. We evaluate 11 architectures across 5 TCGA cancer types, totaling 290 fold-level calibration tests with positive and negative controls, multiple testing correction, and post-hoc recalibration analysis. In Experiment A, all three models fail 1-calibration on a majority of folds, providing direct evidence of miscalibration. Experiment B extends this: 166 of 290 fold-level tests reject calibration after Benjamini-Hochberg correction, though these results reflect Breslow-reconstructed curves rather than native model outputs. Platt scaling recovers much of the lost calibration at the evaluated horizon without sacrificing ranking accuracy.

Our contributions:
\begin{enumerate}
    \item To our knowledge, the first systematic fold-level 1-calibration audit of multimodal WSI-genomics survival architectures. Experiment A directly evaluates native survival outputs from 3 models on TCGA-BRCA, finding all three miscalibrated on a majority of folds. Experiment B extends the audit to 11 architectures across 5 TCGA cancer types via Breslow-reconstructed survival curves; across both experiments, 166 of 290 fold-level tests reject calibration at the median event time after multiple testing correction.
    \item Fusion method is strongly associated with calibration quality. Gating-based fusion is consistently associated with better calibration; bilinear and concatenation fusion are not.
    \item Platt scaling recovers much of the lost calibration at the evaluated time horizon without affecting discrimination.
    \item A validated pipeline with positive and negative controls, establishing a template for future audits.
\end{enumerate}

\subsection*{Generalizable Insights about Machine Learning in the Context of Healthcare}

Three insights extend beyond survival prediction. First, discrimination and calibration are independent properties that must be evaluated separately - a model that ranks well can still produce unreliable probabilities, and the clinical consequences differ. Second, architectural choices appear to carry unintended calibration costs. Fusion methods designed to maximize ranking accuracy are associated with distorted probability estimates; simpler mechanisms may be preferable when probability quality matters. Third, evaluation norms shape what gets optimized. A single ranking metric has created a blind spot across an entire subfield. Adopting calibration evaluation would likely drive improvements in both architecture design and training methodology.

\section{Related Work}

\subsection{Multimodal Survival Prediction}

The integration of WSIs and genomic data for survival prediction has advanced rapidly. MCAT \citep{chen2021multimodal} introduced cross-attention between genomic feature groups and WSI patch embeddings. SurvPath \citep{jaume2024modeling} tokenizes transcriptomics into biological pathway representations and models dense cross-modal interactions via a multimodal Transformer. MMP \citep{song2024multimodal} compresses $>$10,000 WSI patch tokens into 16 prototypes via Gaussian mixture models before fusion. Related models include PORPOISE \citep{chen2022pan}, MOTCat \citep{xu2023multimodal}, and CMTA \citep{zhou2023cmta}. Most use negative log-likelihood discrete survival loss with 4 quantile time bins; MMP uses the Cox partial likelihood loss. Nearly all evaluate exclusively via C-index. MADSurv \citep{zhang2025madsurv} is a recent exception, reporting Brier scores for multimodal cancer survival prediction, though without formal calibration hypothesis testing.

A systematic review of 48 studies combining WSIs and high-throughput omics for survival prediction found that all showed unclear or high risk of bias, with limited external validation and little focus on clinical utility \citep{jennings2025systematic}. None reported formal calibration evaluation.

\subsection{Calibration Evaluation in Survival Analysis}

\citet{haider2020effective} introduced two calibration metrics for individual survival distributions. D-calibration tests whether the probability integral transform of observed times under predicted distributions is uniformly distributed. 1-calibration evaluates whether predicted survival probabilities match observed event rates at a specific time horizon, analogous to the Hosmer-Lemeshow goodness-of-fit test. Both are implemented in SurvivalEVAL \citep{qi2023survivaleval}. \citet{goldstein2020xcal} proposed X-CAL, a differentiable calibration objective for training.

\citet{lillelund2025stop} surveyed 2023-2025 survival analysis publications and found calibration metrics used an order of magnitude less than concordance metrics, recommending evaluation across discrimination, accuracy, and calibration. The A-calibration framework \citep{simonsen2025acalibration} showed that D-calibration becomes conservative under heavy censoring - a relevant limitation for TCGA cohorts with 40-85\% censoring rates.

Two recent benchmarks evaluate calibration for multi-omics survival models on tabular data. SurvBoard \citep{wissel2025survboard} benchmarks 12 models across 28 datasets using C-index, IBS, and D-calibration, finding that statistical models outperform deep learning in calibration. \citet{tran2025comprehensive} evaluate 20 methods across 17 TCGA datasets. Neither includes WSI-based models. VLSA \citep{liu2025vlsa} is the first WSI survival paper to report D-calibration, but only for vision-only and vision-language baselines - not multimodal WSI-genomics models.

\subsection{Uncertainty Quantification for Deep Survival Models}

This field is nascent. NeuralSurv \citep{monod2025neuralsurv} is the first deep survival model with Bayesian uncertainty quantification, applied to tabular data only. SurvUnc (KDD, 2025) is the first model-agnostic post-hoc UQ framework for survival. M2EF-NNs \citep{luo2025m2ef} applies Dempster-Shafer theory to multimodal cancer survival but treats it as discretized classification and lacks calibration evaluation. \citet{morrill2025experts} propose mixture-of-experts architectures targeting calibration alongside discrimination in discrete-time survival, on tabular data. To our knowledge, no existing work conducts a systematic fold-level 1-calibration audit with post-hoc recalibration analysis for multimodal WSI-genomics survival models.

\section{Methods}

\subsection{Models Evaluated}

We conduct two complementary experiments.

\textbf{Experiment A} evaluates three modern multimodal survival models on TCGA-BRCA, trained by us using a consistent protocol. SurvPath uses pathway-level tokenization of transcriptomics (50 Hallmark gene sets) with cross-attention to WSI patch embeddings. MCAT uses 6 functional gene groups with genomic-guided cross-attention over WSI patches. MMP compresses WSI patches into 16 morphological prototypes via Gaussian mixture models before co-attention fusion with 50 pathway prototypes. All use UNI2-h (a subsequent release of UNI; \citealp{chen2024uni}) 1,536-dimensional patch embeddings extracted at 20$\times$ magnification (256$\times$256 patches) and are trained for 20 epochs with 5-fold stratified cross-validation (seed 42) on a single CPU. All three use NLL survival loss over 4 discrete time bins (our MMP implementation uses NLL for consistency with SurvPath and MCAT, though the original MMP paper \citep{song2024multimodal} uses Cox partial likelihood loss; our Experiment A results for MMP therefore reflect NLL-trained behavior). The TCGA-BRCA cohort comprises 1,004 patients after intersecting RNA-seq, WSI features, and survival labels: 149 events, 85.2\% censoring.

\textbf{Experiment B} evaluates 11 architectures using pre-computed results from \citet{chen2021multimodal} across 5 TCGA cancer types: BLCA ($n$$\approx$375), BRCA ($n$$\approx$960), GBMLGG ($n$$\approx$570), LUAD ($n$$\approx$455), UCEC ($n$$\approx$480). Architectures include MCAT; three fusion variants each of attention MIL (AMIL), DeepSets (DS), and multi-input FCN (MIFCN), using gating (sig), bilinear, and concatenation fusion; plus a genomics-only SNN baseline. All use UNI v1 1,024-dimensional features and the same 5-fold CV protocol.

\subsection{Calibration Evaluation Protocol}

\textbf{Per-fold evaluation.} All metrics are computed independently per fold. Pooling validation predictions across folds conflates patients evaluated against different training distributions.

\textbf{Survival curve construction.} Calibration evaluation requires full predicted survival distributions $S(t|x)$, not scalar risk scores. For SurvPath (which outputs 4-bin hazard logits): sigmoid activation, cumulative product, interpolation to 20 time points using training-fold quartile bin edges, monotonicity enforced via cumulative minimum. For all other models (scalar risk scores): Breslow estimation with Kaplan-Meier baseline fit on the training fold, shifted as $S(t|x) = S_0(t)^{\exp(\text{risk} - \text{median\_risk})}$.

\textbf{Metrics.} The concordance index (C-index) measures discrimination. 1-calibration, via SurvivalEVAL's \texttt{one\_calibration()}, is an IPCW-weighted Hosmer-Lemeshow test at the median event time among validation-set events per fold. The Integrated Brier Score (IBS) integrates squared prediction error over time, IPCW-weighted.

\textbf{D-calibration exclusion.} All models produce 4-bin discrete survival curves with 4-5 distinct probability values. D-calibration requires continuous distributions. Our negative control confirms the test lacks power here: shuffled predictions pass D-calibration ($p = 0.80$). We exclude it.

\subsection{Controls}

Three controls validate the pipeline.

\textit{Positive control.} An $L_2$-regularized Cox proportional hazards model (penalizer = 0.1, top 20 variable genes) on the same BRCA data and splits. Cox-PH should pass calibration. It does not reject the null: 1-cal $p = 0.08$ (non-significant at $\alpha = 0.05$, though not far above the threshold). The marginal $p$-value suggests the test is operating near its power limit for this sample size ($\sim$30 events per fold); borderline rejections in the deep learning models should be interpreted with this context in mind.

\textit{Negative control.} SurvPath fold 0 predictions randomly permuted across patients. Should fail. Does: 1-cal $p = 0.0001$, IBS degrades from 0.158 to 0.175.

\textit{Breslow validation.} Our KM-shift construction vs.\ lifelines' \texttt{predict\_survival\_function()} \citep{davidsonpilon2019lifelines} for Cox-PH. The two produce 1-cal $p$-values of 0.079 and 0.078. The construction method does not introduce artifacts.

\subsection{Post-hoc Recalibration}

For each fold $k$, Platt scaling \citep{platt1999probabilistic} fits a logistic regression on predicted survival probabilities from all other folds' validation sets, mapping predicted $S(t_{\text{eval}})$ to binary outcomes (event before $t_{\text{eval}}$). The fitted scaler is applied to fold $k$'s full survival curves at all time points, with monotonicity enforced. No data leakage: the recalibrator never sees the patients it is evaluated on. Recalibrated curves are evaluated using the same SurvivalEVAL \texttt{one\_calibration()} as the main audit. We also test isotonic regression.

\subsection{Multiple Testing Correction}

290 fold-level hypothesis tests (275 Experiment B, 15 Experiment A). Benjamini-Hochberg at FDR = 0.05.

\section{Results}

\subsection{Modern Models on TCGA-BRCA (Experiment A)}

All three models fail 1-calibration on a majority of folds (Table~\ref{tab:perfold}). MCAT is the worst: $p < 0.001$ on every fold, all surviving BH correction. SurvPath rejects on all five folds at raw $\alpha = 0.05$ ($p$-values 0.000--0.033), with four surviving BH correction (fold 1 at $p = 0.033$ does not). MMP passes on one fold ($p = 0.233$); of the four raw rejections, three survive BH correction.

\begin{table}[t]
\centering
\caption{Per-fold 1-calibration $p$-values on TCGA-BRCA (Experiment A). Raw $p$-values shown; values below 0.05 reject at the nominal level. Two borderline cases (SurvPath fold 1, MMP fold 1) do not survive Benjamini-Hochberg correction across all 290 tests.}
\label{tab:perfold}
\begin{tabular}{lcccccc}
\toprule
Model & Fold 0 & Fold 1 & Fold 2 & Fold 3 & Fold 4 & Fails \\
\midrule
SurvPath & 0.018 & 0.033 & 0.000 & 0.000 & 0.001 & 5/5 \\
MCAT     & 0.000 & 0.000 & 0.000 & 0.000 & 0.000 & 5/5 \\
MMP      & 0.233 & 0.037 & 0.012 & 0.000 & 0.000 & 4/5 \\
\bottomrule
\end{tabular}
\end{table}

SurvPath achieves the best C-index ($0.669 \pm 0.063$) but is no more calibrated than MCAT ($0.645 \pm 0.057$). MMP has the lowest C-index ($0.598 \pm 0.081$) and fewest calibration failures (3/5 after BH correction). MMP's IBS excludes fold 4 ($0.120 \pm 0.032$; Table~\ref{tab:summary}) because 8/30 events fell beyond the survival curve's time horizon, producing an anomalous IBS of 3.81.

\begin{table}[t]
\centering
\caption{Summary metrics for Experiment A (TCGA-BRCA). Mean $\pm$ std across 5 folds. 1-cal fails after Benjamini-Hochberg correction (FDR = 0.05) across all 290 tests.}
\label{tab:summary}
\begin{tabular}{lccc}
\toprule
Model & C-index & 1-cal fails (BH) & IBS \\
\midrule
SurvPath & $0.669 \pm 0.063$ & 4/5 & $0.223 \pm 0.117$ \\
MCAT     & $0.645 \pm 0.057$ & 5/5 & $0.311 \pm 0.229$ \\
MMP      & $0.598 \pm 0.081$ & 3/5 & $0.120 \pm 0.032$\textsuperscript{*} \\
\bottomrule
\multicolumn{4}{l}{\footnotesize \textsuperscript{*}MMP fold 4 IBS (3.81) excluded: 8/30 events fall beyond the survival curve's time horizon.}
\end{tabular}
\end{table}

\begin{figure}[t]
\centering
\includegraphics[width=\textwidth]{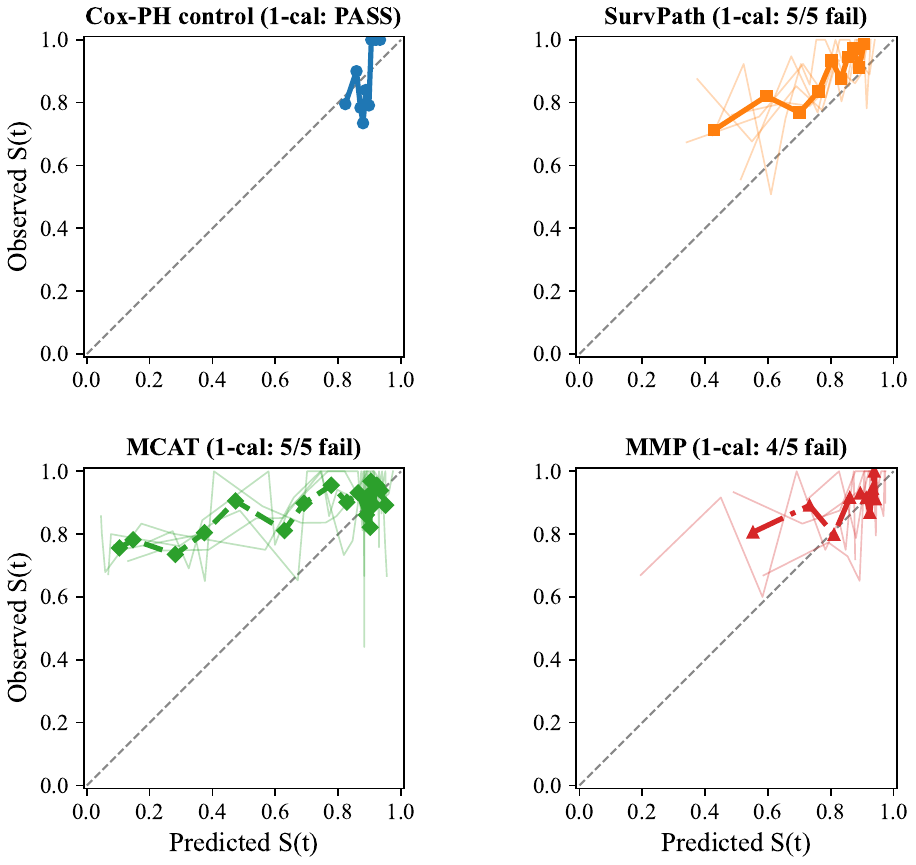}
\caption{Calibration curves on TCGA-BRCA at median event time ($\sim$42 months). Each subplot shows one model. Thick lines: mean across folds. Thin lines: individual folds. Dashed diagonal: perfect calibration. Cox-PH (positive control) tracks the diagonal; deep models deviate systematically.}
\label{fig:brca_cal}
\end{figure}

Figure~\ref{fig:brca_cal} shows calibration curves. Cox-PH tracks the diagonal. SurvPath and MCAT lie above it: for patients predicted to have moderate survival probability (0.4-0.7), actual survival is higher. These models overestimate risk in the middle range. MMP shows a compressed prediction range (0.65-1.0), with less visible but still significant deviation.

\subsection{Multi-Cancer, Multi-Architecture Audit (Experiment B)}

\begin{figure}[t]
\centering
\includegraphics[width=0.75\textwidth]{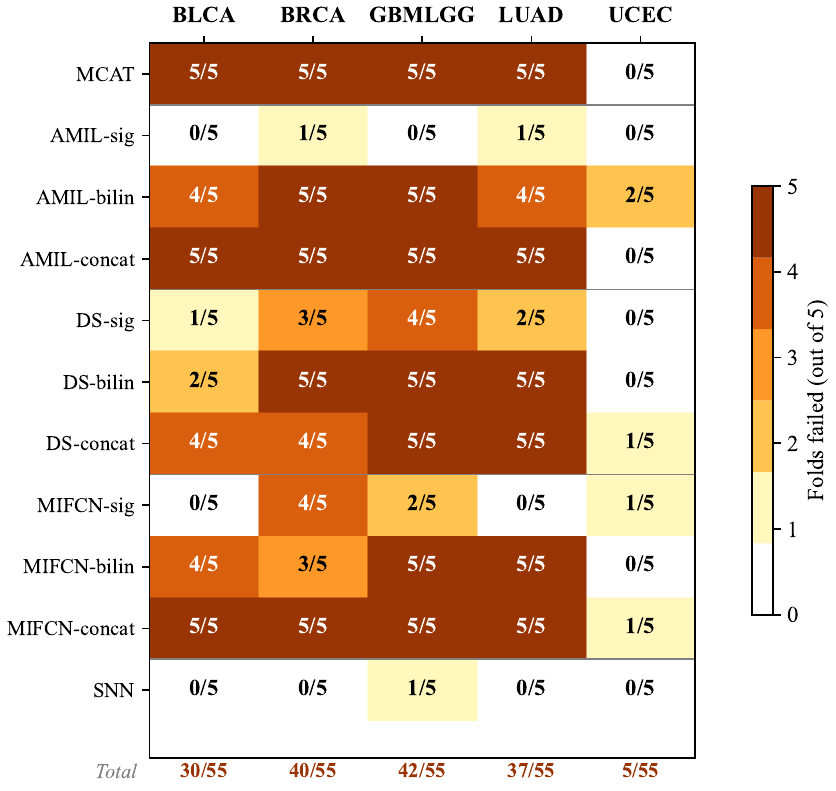}
\caption{1-calibration failure rates across 11 architectures and 5 TCGA cancer types. Each cell shows folds failed out of 5. Darker = more miscalibrated. GBMLGG column is almost entirely dark despite highest discrimination. Gating (sig) rows are consistently lighter than bilinear/concat rows within each family. UCEC is underpowered ($\sim$15 events/fold).}
\label{fig:heatmap}
\end{figure}

Figure~\ref{fig:heatmap} summarizes the broader audit. Of 290 fold-level 1-calibration tests at the median event time (275 from Experiment B's Breslow-reconstructed curves, 15 from Experiment A's NLL-based outputs), 166 survive Benjamini-Hochberg correction at FDR = 0.05. As noted above, Experiment B evaluates reconstructed survival curves, not native model outputs; these results should be interpreted accordingly.

\textbf{Cancer type matters.} GBMLGG has the most severe miscalibration (42/55 failures) despite the highest discrimination (C-index 0.817 for MCAT). BRCA and LUAD show intermediate rates (40/55 and 37/55). BLCA is moderate (30/55). UCEC shows only 5/55 failures, but with $\sim$15 events per validation fold, the Hosmer-Lemeshow test has limited power. This result is ambiguous: it may reflect adequate calibration on this cancer type, or simply insufficient statistical power to detect miscalibration.

\textbf{Fusion method is strongly associated with calibration.} Within the same base architecture, gating is consistently associated with better calibration while bilinear and concatenation are not:

\begin{itemize}
    \item AMIL: gating 2/25 failures, bilinear 20/25, concat 20/25
    \item DS: gating 10/25, bilinear 17/25, concat 19/25
    \item MIFCN: gating 7/25, bilinear 17/25, concat 21/25
\end{itemize}

One possible explanation: gating applies a learned multiplicative mask with fewer free parameters, while bilinear and concatenation create higher-dimensional interaction spaces with more room to overfit the training distribution's probability structure.

The genomics-only SNN baseline fails on 1/25 folds. The multimodal fusion mechanism is more strongly implicated in miscalibration than the individual modalities alone.

\textbf{Important caveat for Experiment B.} Experiment B evaluates Breslow-reconstructed survival curves from scalar risk scores, not native model outputs (Section~3.2). Part of the observed miscalibration may reflect the reconstruction step; see Discussion for a full treatment of this limitation.

\textbf{Discrimination and calibration pull in opposite directions.} No model achieves both high discrimination (C-index $>$ 0.65) and consistent calibration ($\leq$1 fold failing). Models that learn useful representations produce unreliable probabilities. Models with reliable probabilities (SNN, AMIL-sig) have limited discriminative power.

\begin{figure}[t]
\centering
\includegraphics[width=\textwidth]{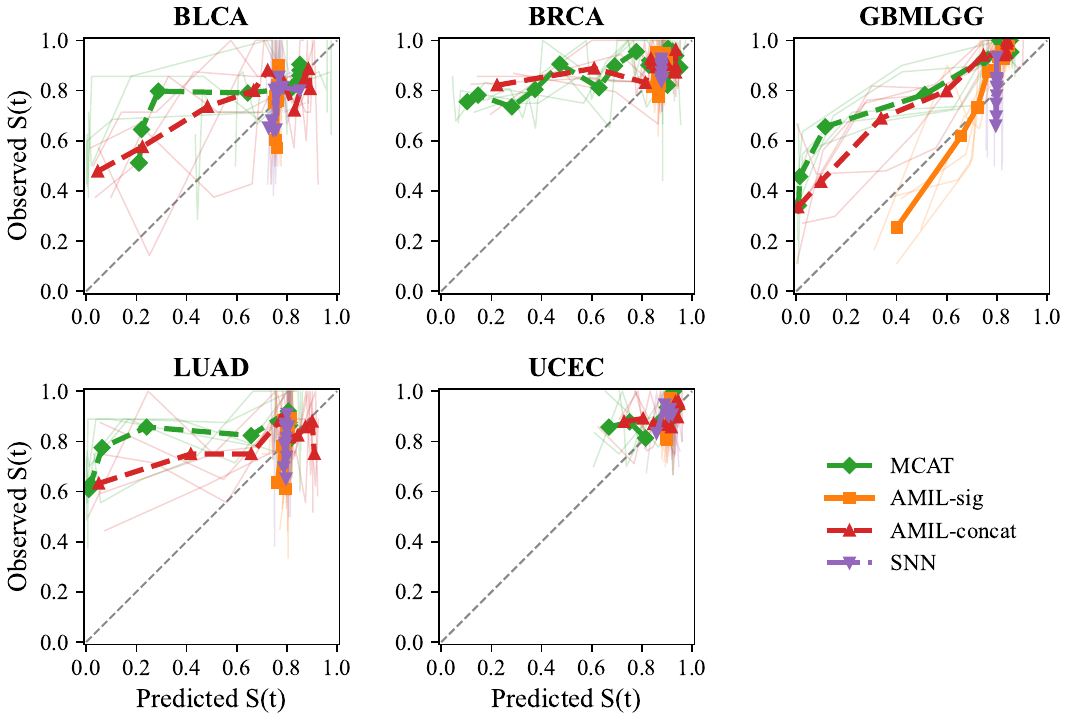}
\caption{Calibration curves across 5 TCGA cancer types for 4 representative architectures. GBMLGG shows the most dramatic deviation from the diagonal. UCEC shows all models clustered near the diagonal, consistent with underpowered testing.}
\label{fig:multicancer}
\end{figure}

\subsection{Post-hoc Recalibration}

\begin{figure}[t]
\centering
\includegraphics[width=\textwidth]{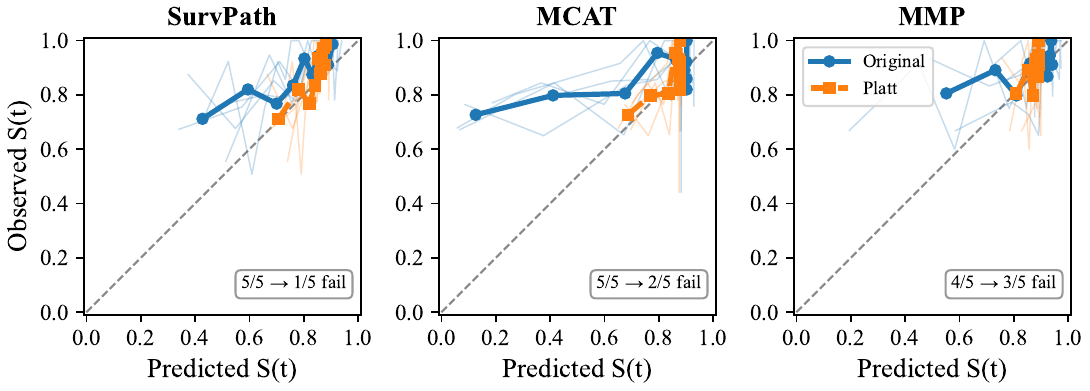}
\caption{Calibration curves before (blue) and after (orange) Platt scaling. Annotations show improvement in fold-level failure rates. MCAT shows the clearest visual improvement.}
\label{fig:recal}
\end{figure}

Platt scaling improves calibration for all three Experiment A models (Table~\ref{tab:recal}, Figure~\ref{fig:recal}).

\begin{table}[t]
\centering
\caption{Recalibration results (TCGA-BRCA, Experiment A). Evaluated using SurvivalEVAL \texttt{one\_calibration()} for consistency with the main audit.}
\label{tab:recal}
\begin{tabular}{lcccc}
\toprule
Model & Original & Platt & Original IBS & Platt IBS \\
\midrule
SurvPath & 5/5 fail & 1/5 fail & 0.223 & 0.216 \\
MCAT     & 5/5 fail & 2/5 fail & 0.311 & 0.219 \\
MMP      & 4/5 fail & 3/5 fail & 0.120\textsuperscript{*} & 0.127\textsuperscript{*} \\
\bottomrule
\multicolumn{5}{l}{\footnotesize \textsuperscript{*}MMP fold 4 excluded (same time-horizon issue as Table~\ref{tab:summary}).}
\end{tabular}
\end{table}

C-index is unchanged in all cases. Platt scaling is monotonic; ranking is preserved.

Isotonic regression does not help (4/5 folds failing for all three models), likely due to overfitting with $\sim$360 training points. The models rank patients well; the mapping from risk to probability appears to be the issue - directly for Experiment A's native outputs, and as observed under Breslow reconstruction for Experiment B - and at the evaluated horizon a two-parameter correction largely recovers calibration.

\section{Discussion}

\textbf{Two levels of evidence.} Experiment A provides the more direct evidence: all three models are trained with NLL discrete survival loss and produce survival curves through the model's own discretization and hazard-to-survival conversion, without post-hoc Breslow reconstruction. SurvPath natively outputs per-bin hazard logits; MCAT and MMP use the same NLL-based pipeline. All three fail calibration on a majority of folds. Experiment B provides broader but more indirect evidence: the survival curves are reconstructed from scalar risk scores via Breslow estimation, so the observed miscalibration may partly reflect the reconstruction step. Nonetheless, the pattern across 11 architectures and 5 cancer types is strikingly consistent, and Breslow estimation is the standard post-hoc method that would be used in any deployment of these models. Notably, calibration failures on BRCA appear under both UNI v1 features (Experiment B) and UNI2-h features (Experiment A), suggesting the feature extractor alone does not explain the problem -though the two experiments differ in too many variables for this to constitute an ablation.

\textbf{Why are these models miscalibrated?} The training objective is the most likely contributor. NLL discrete survival loss rewards correct ranking but does not penalize miscalibrated probabilities; the 4-bin discretization compounds the issue. The pattern is consistent across architectures regardless of other design choices, suggesting the problem is common to discrimination-focused objectives. Calibration-aware losses exist -X-CAL \citep{goldstein2020xcal} adds a differentiable calibration penalty -but none have been tested on multimodal WSI-genomics models. That Platt scaling largely fixes the problem with two parameters hints that even a modest calibration term in the loss could help.

Gating-based fusion is consistently associated with better calibration across all three architecture families and five cancer types. We hypothesize this relates to expressiveness: gating applies a constrained multiplicative mask, while bilinear and concatenation fusion create higher-dimensional interaction spaces that may allow greater distortion of the probability mapping. This remains a hypothesis; the current design cannot establish a causal mechanism.

Consider GBMLGG. MCAT achieves C-index 0.817 yet fails 1-calibration on all five folds. The model ranks glioma patients well, but the survival probabilities it assigns are unreliable. Where those probabilities might influence treatment decisions, the distinction between good ranking and poor calibration is clinically consequential.

\textbf{Recalibration helps at the evaluated horizon.} That is the encouraging part. Platt scaling is a two-parameter logistic transformation. Its effectiveness at the median event time tells us the miscalibration there is systematic, not random. The models have learned meaningful representations of survival risk; at least at the median event time, they map those representations to the wrong probability scale. For Experiment A models with native survival outputs, this is direct evidence of a correctable systematic bias. For Experiment B, the same pattern holds under Breslow reconstruction, though the reconstruction step may contribute. In both cases, a monotonic correction recovers much of the lost calibration at the evaluated horizon. Since the scaler is fit at a single time point and applied globally, improvement at other horizons is plausible but unverified. The practical implication: deployment of these models should consider including a post-hoc calibration step, ideally evaluated at multiple clinically relevant horizons. The deeper implication: improving calibration may not require fundamental architectural changes.

Three recommendations follow: (1) report calibration metrics alongside C-index in future survival model publications; (2) consider post-hoc recalibration before clinical use of existing models; (3) investigate calibration-aware training objectives for multimodal survival models.

These recommendations align with the FDA's January 2025 draft guidance on AI-enabled device software functions \citep{fda2025aidsf}, which discusses calibration and uncertainty among considerations for performance evaluation of higher-risk applications.

\paragraph{Limitations}

Experiment B's Breslow-based survival curve construction assumes proportional hazards; deep models may violate this, meaning part of the observed miscalibration could reflect the reconstruction method rather than the models themselves. The Breslow validation control (Appendix~C) only confirms the construction is artifact-free for Cox-PH, which satisfies the proportional-hazards assumption by definition; it does not validate the construction for deep models that may violate PH. 1-calibration evaluates at one time point (median event time); miscalibration at other clinically relevant horizons (e.g., 1-year, 5-year) is not assessed, and results should be interpreted as calibration at this specific horizon rather than a general calibration verdict. Platt scaling is similarly fit and evaluated at a single horizon. Our evaluation relies on binary hypothesis testing (reject/not reject); continuous calibration error measures (e.g., mean absolute calibration error, calibration slope) would provide more granular assessment of miscalibration magnitude. IBS combines calibration and discrimination and should not be interpreted as a pure calibration measure. Fold-level 1-calibration $p$-values depend on effective event counts, reduced by high censoring (40--85\%) and 5-fold splitting; UCEC ($\sim$15 events/fold) may be underpowered. Our power analysis shows the Hosmer-Lemeshow test exhibits inflated type I error at these sample sizes (17--21\% rejection rate under the null at $n = 15$--$37$ events, vs.\ the nominal 5\%), meaning some individual rejections may be false positives; however, the observed rejection count (166/290) far exceeds even the inflated null expectation ($\sim$58 rejections at 20\% type I error), so the overall conclusion of widespread miscalibration is robust. IPCW weighting, used by 1-calibration, becomes unstable under heavy censoring; under BRCA's 85\% censoring, IPCW reduces effective sample sizes to 52--68\% of actual, with weight coefficients of variation of 0.69--0.79. All results use TCGA data; generalization across institutions is untested. D-calibration is excluded due to discrete 4-bin outputs. Recalibration was evaluated only on Experiment A. The fusion-calibration association is observational, not causal. Our MMP implementation uses NLL loss rather than the original Cox partial likelihood.

\section{Conclusion}

Across 290 fold-level 1-calibration tests spanning 11 architectures and 5 TCGA cancer types, 166 reject the null of correct calibration at the median event time after Benjamini-Hochberg correction. The pattern is consistent: models optimized for discrimination produce unreliable survival probabilities, gating-based fusion is associated with better calibration than bilinear or concatenation alternatives, and post-hoc Platt scaling recovers much of the lost calibration without affecting ranking accuracy. These findings argue for three practical changes: reporting calibration metrics alongside C-index, applying post-hoc recalibration before clinical deployment, and investigating calibration-aware training objectives for multimodal survival models.

\paragraph{Code and Data Availability}
Code for all experiments, calibration evaluation, and recalibration is available at \url{https://github.com/sajadghawami/calibration-audit-survival}. All experiments use publicly available TCGA data accessed via the Genomic Data Commons.

\bibliography{references}

\appendix
\section*{Appendix A. Cohort Details}

\textbf{TCGA-BRCA (Experiment A):} 1,004 patients (RNA-seq $\cap$ WSI features $\cap$ survival labels). 149 deaths (14.8\%), 855 censored (85.2\%). Median survival among events: 1,274 days ($\sim$42 months). Excluded: survival\_time $<$ 30 days. RNA-seq: GENCODE v36, log1p TPM. WSI: UNI2-h ViT-H/14, 1,536-dim, 256$\times$256 patches at 20$\times$.

\textbf{MCAT ICCV cohorts (Experiment B):} BLCA ($n$$\approx$375), BRCA ($n$$\approx$960), GBMLGG ($n$$\approx$570), LUAD ($n$$\approx$455), UCEC ($n$$\approx$480). All use 5-fold stratified cross-validation from \citet{chen2021multimodal}.

\section*{Appendix B. D-calibration Analysis}

D-calibration tests uniformity of the probability integral transform (PIT). For continuous survival distributions, PIT values should be $\text{Uniform}(0,1)$ under correct calibration. All evaluated models produce 4-bin discrete survival curves with only 4-5 distinct probability values, violating the continuity assumption. Our negative control confirms the test has no power at this resolution: shuffled predictions pass D-calibration ($p = 0.80$), while 1-calibration correctly rejects ($p = 0.0001$).

\section*{Appendix C. Breslow Validation}

For models producing only scalar risk scores, we construct survival curves via Kaplan-Meier baseline estimation on training data, shifted by $S(t|x) = S_0(t)^{\exp(\text{risk} - \text{median\_risk})}$. To validate this approximation, we compared it against lifelines' \texttt{predict\_survival\_function()} for a Cox-PH model. The two methods produce 1-calibration $p$-values of 0.079 and 0.078, confirming no artifacts from the construction method (see Figure~\ref{fig:breslow}).

\noindent
\begin{minipage}{\textwidth}
\centering
\includegraphics[width=0.5\textwidth]{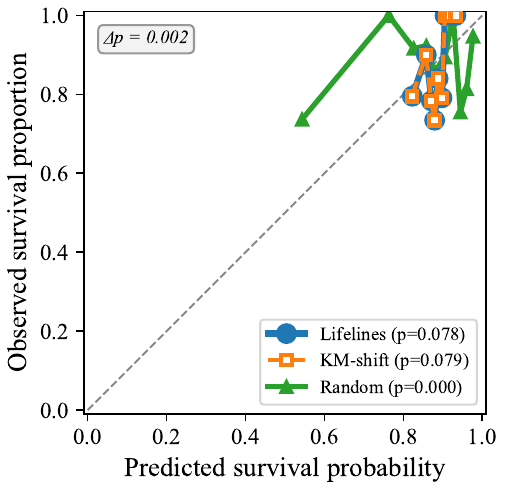}
\captionof{figure}{Breslow validation: lifelines vs.\ KM-shift survival curve construction for Cox-PH. Both produce nearly identical calibration curves ($\Delta p = 0.002$). Random assignment serves as a negative control.}
\label{fig:breslow}
\end{minipage}

\end{document}